\documentclass[10pt,twocolumn,letterpaper]{article}

\usepackage{cvpr}              
\definecolor{cvprblue}{rgb}{0.21,0.49,0.74}
\usepackage[pagebackref,breaklinks,colorlinks,allcolors=cvprblue]{hyperref}

\usepackage{amsmath,amsfonts,bm}

\def\1{\bm{1}}

\def\vtheta{{\bm{\theta}}}

\def\vm{{\bm{m}}}

\def\vx{{\bm{x}}}

\def\vz{{\bm{z}}}

\DeclareMathAlphabet{\mathsfit}{\encodingdefault}{\sfdefault}{m}{sl}
\SetMathAlphabet{\mathsfit}{bold}{\encodingdefault}{\sfdefault}{bx}{n}

\def\gA{{\mathcal{A}}}

\def\gM{{\mathcal{M}}}
\def\gN{{\mathcal{N}}}

\usepackage{booktabs}
\usepackage{pifont}
\newcommand{\cmark}{\ding{51}}  
\newcommand{\xmark}{\ding{55}}  
\usepackage[table]{xcolor}
\definecolor{Gray}{gray}{0.9}

\def\paperID{*****} 
\def\confName{CVPR}
\def\confYear{2026}

\title{Text-to-seed generation: Training-free open-vocabulary seeded semantic segmentation via re-purposing diffusion as text-guided seed generator}

\author{Kumju Jo$^{1}$ \quad
Heesun Jung$^{2}$ \quad,
Sungyong Baik$^{1,2\dagger}$\\
$^{1}$Dept. of Artificial Intelligence, Hanyang University\\
$^{2}$Dept. of Data Science, Hanyang University\\
{\tt\small \{juice0630,jheesun,dsybaik\}@hanyang.ac.kr}\\
}

\begin{document}
\maketitle

\begin{abstract}
Open-vocabulary semantic segmentation (OVSS) aims to segment image regions corresponding to arbitrary text queries. 
Although the Segment Anything Model (SAM) is a powerful foundation model for segmentation, its standalone performance on OVSS remains limited. 
Existing methods therefore often use SAM to refine coarse masks predicted by other models, but this strategy is unreliable when the initial masks are inaccurate. 
In this work, we argue that more reliable segmentation can be achieved by exploiting SAM as a region expansion module guided by accurate object points (i.e., seeds) rather than inaccurate coarse masks. 
Inspired by classical seeded segmentation, we reformulate OVSS as text-guided seed localization followed by seed-based region expansion. 
To realize this idea, we propose Text-to-Seed (T2S), a training-free framework that leverages the text-to-region correspondence of Stable Diffusion to generate attention-based seed points for target categories described by text. 
These sparse seeds are then used as point prompts for SAM to produce full object masks. 
Without task-specific training or additional annotations, T2S achieves strong performance on standard OVSS benchmarks, demonstrating the effectiveness of combining semantic grounding with seed-driven spatial segmentation.

{\small
\noindent\copyright\ 2026. This manuscript version is made available under the CC-BY-NC-ND 4.0 license
\url{https://creativecommons.org/licenses/by-nc-nd/4.0/}
\par}
\end{abstract}

\section{Introduction}
\label{intro}

Open-vocabulary semantic segmentation (OVSS)~\cite{ZS3Net,ZSSeg,OpenSeg} aims to handle open-set categories in arbitrary texts, generalizing beyond predefined categories.
OVSS task demands for the generalization capability in modeling correspondences between visual and textual representations.

To tackle such challenging task, few works have attempted to employ foundation models for segmentation (particularly, Segment Anything Model (SAM)~\cite{SAM}).
While SAM exhibits strong capability in learning spatially coherent visual representations, SAM faces challenges with text prompts and semantic understanding~\cite{SAMantic}.
Thus, recent works either fine-tune SAM with vision-language models or use SAM to refine coarse segmentation masks or bounding boxes~\cite{CaR, SAMCLIP, RIM}.
However, fine-tuning is susceptible to risks of suboptimal performance on unseen categories, while refinement approaches are limited by the quality of coarse segmentation masks or bounding boxes.

Another line of works~\cite{FreeDA,RIM} focuses on employing text-to-image generative model, namely Stable Diffusion (SD)~\cite{stable-diffusion}, for its capability to spatially align visual features and textual features.
SD is used to either directly estimate segmentation masks from attention maps~\cite{FOSSIL,DiffSegmenter,OVdiff,EmerDiff} or generate synthetic images to prepare semantic features~\cite{FreeDA,RIM,FreeSeg-Diff}. 
However, since SD is originally designed for image generation, these approaches are often faced with the difficulty of generating high-quality masks~\cite{DiffuMask} or recognizing multiple subjects in an image.

\begin{figure}
    \centering
    \includegraphics[width=1\columnwidth]{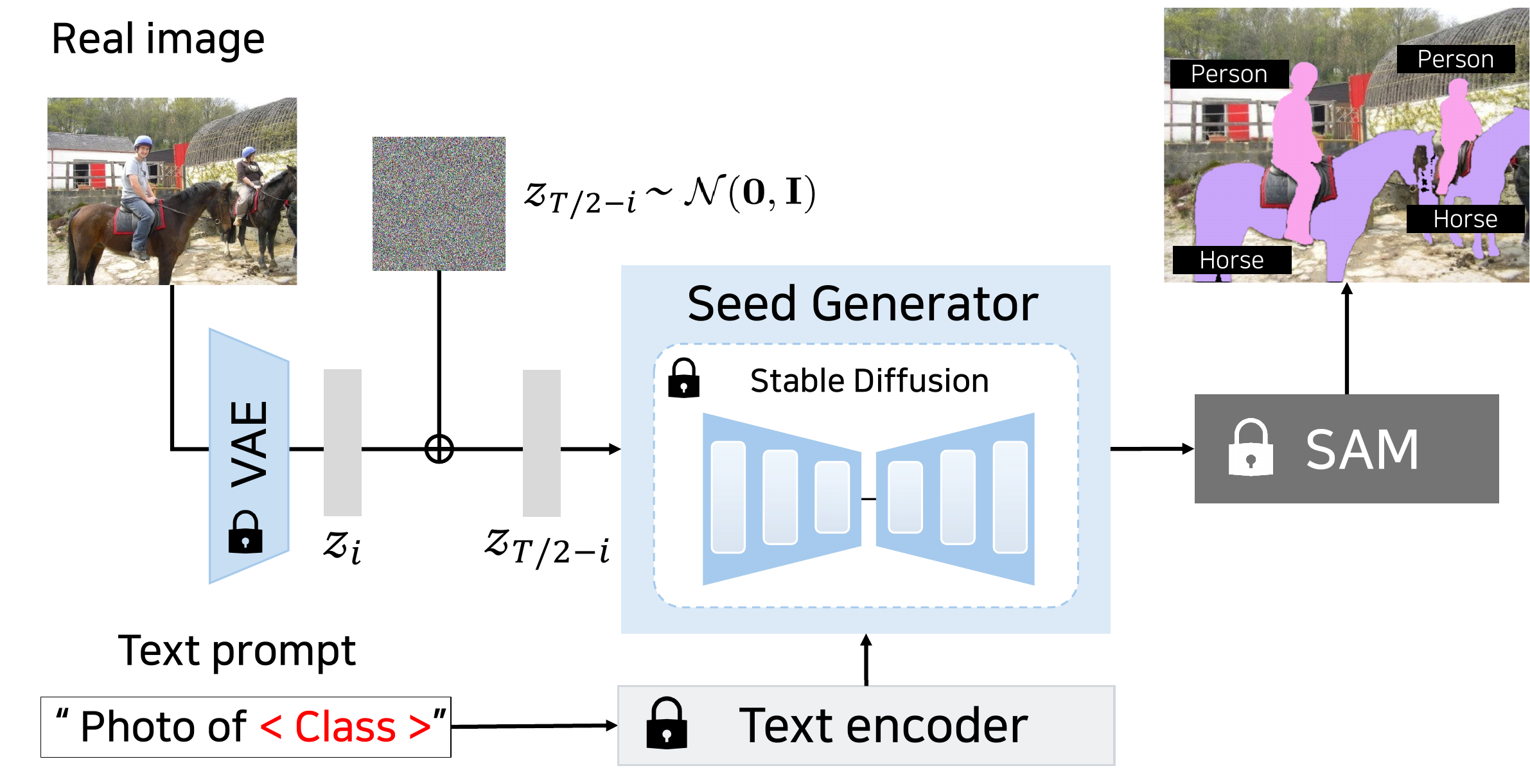}
    \caption{Overview of our framework, dubbed T2S. 
    We extract seeds from attention maps of Stable Diffusion.
    Then, seeds are fed into SAM to perform region expansion.
    }
    \label{fig:architecture}
\end{figure}

In this work, we tackle OVSS from the perspective of seeded segmentation~\cite{seeded_segmentation}, which comprises two stages: seed initialization to localize an object and region expansion to find similar features and segment the whole object.
From this perspective, we draw connections between seeded segmentation and the capabilities of SAM and SD.
Notably, the spatial text-image alignment capability of SD can be used for seed initialization; while, the spatial coherence modeling proficiency of SAM for region expansion.
Specifically, we leverage the attention maps in SD to localize a queried category with initial seed.
This initial seed is then passed to SAM as a point prompt to obtain a high-quality mask.
The overall framework, dubbed Text-to-seed (T2S), is illustrated in Fig.~\ref{fig:architecture}.

Experimental results highlight the outstanding performance of our proposed framework across various datasets.
We note that our framework is a plug-and-play framework that is designed for SAM and SD to be seamlessly incorporated in an off-the-shelf manner, without any additional training.
The strong performance and flexibility of T2S corroborate our motivation and idea that draws connections with classical seeded segmentation, re-purposing SD for seed generation and SAM for region expansion.    
\section{Related Works}
\label{sec:related}

Traditional semantic segmentation tasks have focused on assigning each pixel with  pre-defined categories~\cite{coco, Ade20k, Cityscapes, pascalVOC}.
Despite rapid and notable progresses with deep learning, the reliance on predefined categories has limited its generalization capability.
To take a step further and generalize beyond such assumption, open-vocabulary semantic segmentation (OVSS) aims to assign each pixel with any given category prompted by arbitrary text queries.

To tackle OVSS, many works have employed Vision-Language Models (VLMs)~\cite{CLIP,ALIGN,FLIP,Eva-clip,CoCa,Lit,sigmoid_VLM}, especially CLIP, trained on a large-scale dataset of text-image pairs.
However, CLIP is originally trained for modeling image-level correspondences with textual features, exhibiting poor spatial coherence and thus providing poor mask quality.
Many works have tried to endow CLIP with spatial coherence modeling via fine-tuning~\cite{ZSSeg, GroupViT, PACL, OVSegmentor, TCL}.
Yet, these works still struggle with achieving high-quality masks for arbitrary categories.
Recent methods~\cite{naclip,cliper1} improve the spatial representation of frozen CLIP, with CLIPer further utilizing diffusion self-attention to capture fine-grained spatial details.
However, these methods mainly improve the spatial consistency of CLIP-derived predictions without introducing an independent text-conditioned localization signal, and thus remain dependent on the semantic quality of the initial CLIP predictions.

Recent studies have shown that text-guided Diffusion Models (DMs), particularly Stable Diffusion (SD)~\cite{stable-diffusion}, not only achieve strong text-to-image generation performance but also encode rich semantic and spatial cues in their latent representations and attention maps~\cite{DAAM,Prompt-to-Prompt,attend_excite}.
Such controllability and representational richness have enabled DMs to be extended to diverse generation-related tasks, including controllable generation through attention guidance~\cite{attend_excite,StructuredDiffusion}, attention-based image editing~\cite{Prompt-to-Prompt,masactrl,PnP}, fair generation with attention modulation~\cite{kim2026training}, personalization using attention maps or attention modules~\cite{vico,degu,SubjectDiffusion}, concept erasure~\cite{egloce,huang2024receler}, and debiasing with editing models~\cite{biasedit}.
Motivated by these advances, several works have investigated the use of DMs for open-vocabulary semantic segmentation (OVSS).
Few works~\cite{FreeDA,RIM,FOSSIL,DiffSegmenter,OVdiff,EmerDiff,FreeSeg-Diff,seediff} have attempted to leverage the generative capability of SD to generate several images for a given category, extracting region features from synthetic images.
The extracted region features are then compared against features of a given real image to perform region-to-region feature matching, bypassing the difficulties of patch-text correspondence modeling.
Meanwhile, other works~\cite{DiffSegmenter,OVdiff,EmerDiff,FreeSeg-Diff} utilize attention maps and of text-to-image correspondence modeling capability of SD to directly obtain semantic masks. 
More recently, ~\cite{nerve1} further combines CLIP cross-attention and Stable Diffusion self-attention with an entropy-guided random walk for mask refinement.
However, direct generation of semantic masks from attention maps have often led to coarse masks.

In parallel, few works~\cite{GroundedSAM,RIM,CaR} have attempted to tackle OVSS with Segment Anything Model (SAM)~\cite{SAM}, a recent vision foundation model for object segmentation.
However, such capability is only applicable to class-agnostic masks, lacking the semantic understanding~\cite{SAMantic}.
Although recent SAM models (i.e., SAM 3)~\cite{sam3} have improved text-prompt handling, their semantic understanding is effectively limited to simple noun phrases. 
This inherent constraint in semantic grounding indicates that these models still fail to independently process arbitrary open-vocabulary queries.
GroundedSAM~\cite{GroundedSAM} introduces a workaround by performing open-vocabulary object detection with Grounding DINO and then feeding a bounding box into SAM to perform segmentation.
Meanwhile, few works~\cite{RIM,CaR} have attempted to refine their coarse masks with masks produced by SAM in post-processing.
However, initial poor-quality masks can limit the extent of refinement SAM masks can give during post-processing.

Recent methods~\cite{shi2024trident,corrclip}  further integrate SAM into CLIP-based dense inference using SAM-derived correlations or region masks.
These methods perform spatial refinement using CLIP-based initial predictions or class-agnostic region priors from SAM; therefore, their final performance can still depend on the quality of the initial semantic localization and region proposals.
On the other hand, our proposed framework, named Text-to-Seed (T2S), better exploits the spatial coherence capabilities of SAM by drawing connections with seeded segmentation.
In particular, \textcolor{black}{T2S re-purposes the diffusion model as a text-guided seed generator to produce point-level accurate localization (seeds); based on these, T2S leverages SAM for region expansion rather than for conventional coarse mask or bounding box refinement}.

\section{Background}
\label{sec:preliminaries}
\subsection{Diffusion Model} 
Stable Diffusion (SD) is one of instances of ~\cite{stable-diffusion} a text-guided diffusion model designed to generate an image $\vx$ from random noise $\vz\sim\gN(\mathbf{0},\mathbf{I})$, based on a text embedding $\bm{\tau}(y)$ obtained from an input text prompt \( y \) of length \( P \) through a text encoder $\bm{\tau}$. 
Diffusion models work under the assumption that $\vz$ and $\vx$ are linked by a Markov chain, which incrementally adds noise to $\vx=\vz_0$, transforming it into $\vz=\vz_T$ over a predetermined number of timesteps \( T \):
\begin{equation}
\label{eq:sampling}
q(\vz_t|\vz_{t-1}) = \gN(\vz_t;\sqrt{1-\beta_t}\vz_{t-1},\beta_t\mathbf{I}) \, ,
\end{equation}
where \( \beta_t \) denotes the noise or variance schedule.

From this perspective, an image $\vx$ can be viewed as a denoised version of $\vz$.
To perform such denoising process to generate $\vx$, a network $\bm{\epsilon}_\vtheta$, parameterized by $\vtheta$, is trained to predict the added noise $\bm{\epsilon}\sim\gN(\mathbf{0}, \mathbf{I})$ at each timestep \( t \), conditioned on the text prompt \( y \), by minimizing the following objective function:
\begin{equation}
\label{eq:sd_loss}
\mathbb{E}_{\vx,\bm{\epsilon}\sim\gN(\mathbf{0},\mathbf{I}),t,\bm{\tau}(y)}\left[\big\| \bm{\epsilon} - \bm{\epsilon}_\vtheta(\vz_t, t, \bm{\tau}(y)) \big\|^2_2 \right].
\end{equation}

\subsection{Aggregation of attention maps} 
The denoising network in Stable Diffusion is composed of residual and transformer blocks. 
Each transformer block includes cross-attention (CA) and self-attention (SA) layers, with a total of \( L = 16 \) transformer blocks. 
At the \( l \)-th transformer layer, the \( l \)-th CA layer merges information from a conditioning text prompt embedding $\bm{\tau}(y)\in\mathbb{R}^{P\times d_\tau}$ and latent features $\vz^l_t\in\mathbb{R}^{H_l\times W_l\times d_z}$ from a previous layer, facilitating image generation aligned with the conditioning text. 
We can calculate the CA map $\mathcal{A}_{l,t}^\text{CA}\in\mathbb{R}^{H_l\times W_l \times P}$ from CA layer, at each timestep \( t \) as follows:
\begin{equation}
\mathcal{A}_{l,t}^\text{CA}=\text{softmax}\left(\frac{Q_{l,t}^{z}\cdot {K_{l}^{\tau}}^\top}{\sqrt{d_l}}\right),
\end{equation}
where \( Q_{l,t}^{z} \) and \( K_{l}^{\tau} \) are the query matrix of $\vz^l_t$ and the key matrix of $\bm{\tau}(y)$, respectively, obtained via linear projections. 
\( d_l \) denotes the latent feature dimension at layer \( l \). The \( l \)-th SA layer, meanwhile, captures spatial similarities within $\vz^l_t$, generating an SA map $\mathcal{A}_{l,t}^\text{SA}\in\mathbb{R}^{H_l\times W_l \times H_l\times W_l}$ by the following:
\begin{equation}
\mathcal{A}_{l,t}^\text{SA}=\text{softmax}\left(\frac{Q_{l,t}^{z}\cdot {K_{l,t}^{z}}^\top}{\sqrt{d_l}}\right),
\end{equation}
where \( Q_{l,t}^{z} \) and \( K_{l,t}^{z} \) are the query and key matrices of $\vz^l_t$.
Stable Diffusion's U-Net architecture, with attention layers in both the encoder and decoder sections, generates attention maps at four different resolutions: \( (H_l, W_l) \in \{ (s_r, s_r) \}_{r=0}^3 \) with \( s_r \in \{8, 16, 32, 64\} \). 

The numerous attention maps generated by Stable Diffusion can be simplified by grouping them by resolution and then averaging and normalizing them to a 0-1 range:
\begin{equation}
\bar{\gA}_{s_r}=\frac{1}{\left|\mathbb{L}_{s_r}\right|\cdot T}\sum_{l\in \mathbb{L}_{s_r}}\sum_{1\le t\le T}\frac{\mathcal{A}_{l,t}}{\max(\mathcal{A}_{l,t})},
\end{equation}
The set $\mathbb{L}_{s_r}$ contains the indices of layers that produce attention maps at resolution \( s_r \), and $\bar{\gA}_{s_r}$ denotes the aggregated attention map for that resolution. 
$\bar{\gA}^\text{CA}_{s_r}$ and $\bar{\gA}^\text{SA}_{s_r}$ represent the aggregated CA and SA maps, respectively. 
\begin{figure*}
    \centering
    \includegraphics[width=1\linewidth]{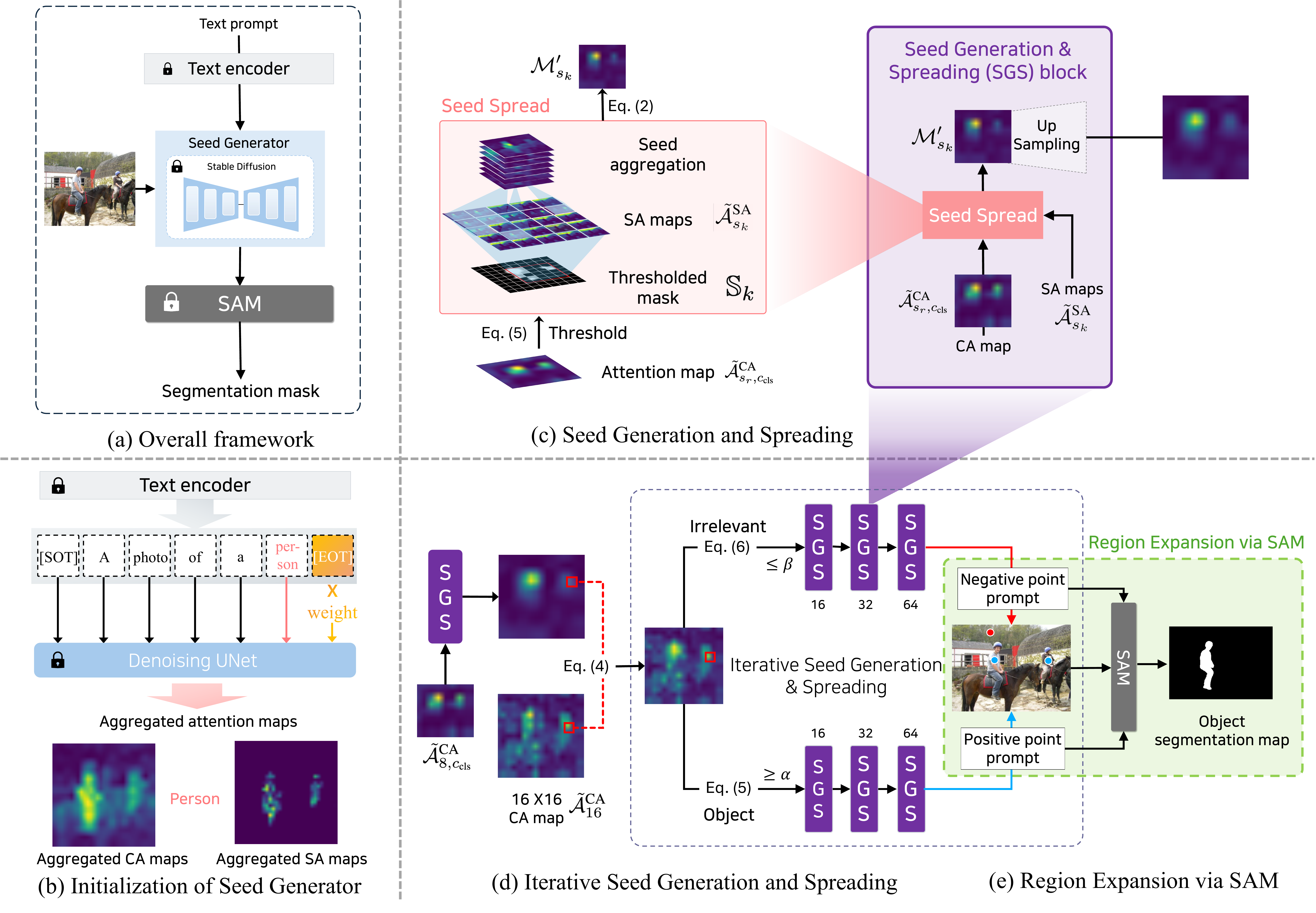}
    \caption{
    (a) \textbf{The simplified overview of our framework}. 
    (b) \textbf{Initialization of seed generator} is performed via Stable Diffusion (SD) with EoT token weighting, followed by aggregating attention maps. 
    Once attention maps are aggregated, SD is no longer needed.
    (c) \textbf{Seed generation and Spreading (SGS) block} extracts and spreads seeds from aggregated attention maps to localize a given category.
    (d) \textbf{Iterative Seed Generation and Spreading process} iteratively applies SGS block that extracts both positive seeds for categories and negative seeds for irrelevant or background, which are iteratively refined to high resolution.
    Since this process is performed on already aggregated attention maps, it can be performed efficiently.
    Then, (e) \textbf{Region Expansion process via SAM} is performed by feeding positive seeds and negative seeds into SAM as point prompts to perform region expansion to get a final mask.
    }
     \label{fig:enter-label}
\end{figure*}

\section{Proposed Method}
Motivated by classical seeded segmentation, our proposed method, \textbf{Text-to-Seed (T2S)} with the overview illustrated in Fig.~\ref{fig:enter-label}(a), performs iterative seed initialization, region expansion, and classification.
Overall, T2S consists of four stages.
The first stage is dedicated to improving text-spatial-visual-feature alignment information from attention maps of SD (Sec.~\ref{sec:initialization}, Fig.~\ref{fig:enter-label}(b)), serving as seed generator in the following section.
The set of obtained attention maps, acting as seed generator, is then used to extract seeds that locate a queried category (Sec.~\ref{sec:seed_generator}, Fig.~\ref{fig:enter-label}(c)).
Then, seeds are refined to higher resolution via iterative seed generation and spreading procedure (Sec.~\ref{sec:ses}, Fig.~\ref{fig:enter-label}(d)).
The extracted seeds are then fed into Segment Anything Model (SAM) as point prompts to perform region expansion (Sec.~\ref{sec:region_expansion}, Fig.~\ref{fig:enter-label}(e)).

\subsection{Initialization of seed generator:\\ enhancing attention maps of stable diffusion}
\label{sec:initialization}
In seeded segmentation, precise mask generation hinges on an accurate seed initialization, locating a queried category.
In this work, we aim to obtain an accurate seed initialization from Stable Diffusion (SD)~\cite{stable-diffusion}.
SD is a diffusion model with U-Net architecture composed of self-attention (SA) layers and cross-attention (CA) layers.
Guided by text prompts, SD learns to generate images ($t=0$) from random Gaussian noise ($t=T$), where $t$ is a denoising step and $T$ is a predefined number of denoising steps. 

To extract accurate initial seeds from SD, we first aim to amplify the signal from a target category $c$ in a given text prompt.
We take motivation from recent analysis on DMs that the attention maps corresponding to End-of-Text ([EOT]) embeddings contain semantic information of objects present in the same text~\cite{EoT}.
Upon the motivation, we up-weight [EOT] embeddings by a factor of two to strengthen the signal of a target category in the attention maps.

\begin{figure}
    \centering
    \includegraphics[width=1\linewidth]{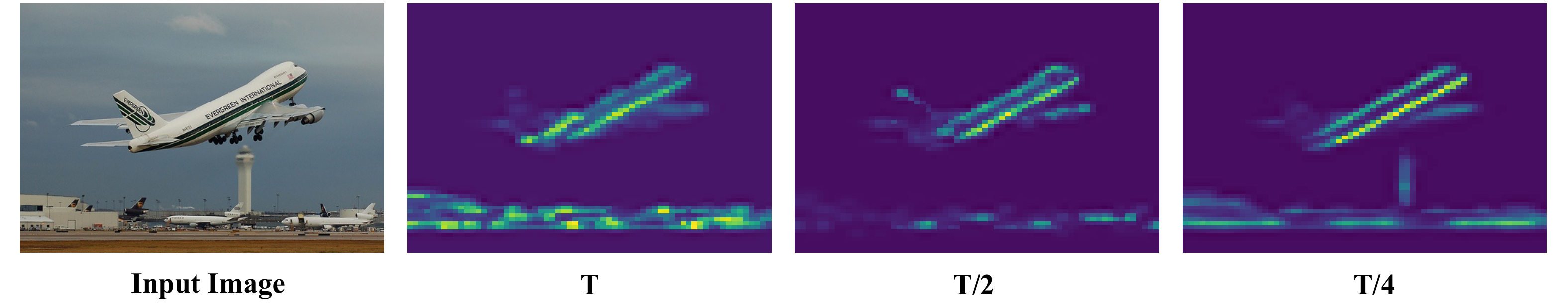}
    \caption{
    Visualization of how aggregated cross-attention (ACA) maps change with different starting denoising steps.
    }
    \label{fig:timestep}
\end{figure}

\begin{figure}
    \centering
    \includegraphics[width=1\linewidth]{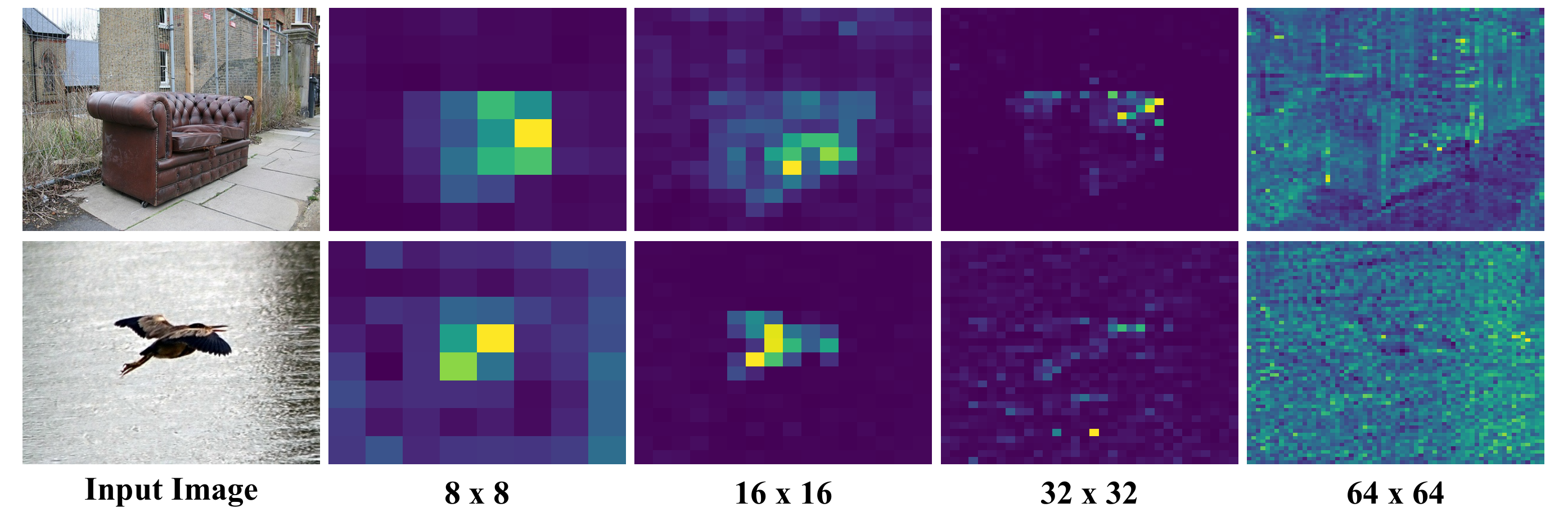}
    \caption{
    Visualization of cross-attention aggregation (ACA) maps at different resolutions in Stable Diffusion.
    }
    \label{fig:ca_from_res}
\end{figure}

Next, we decide which diffusion denoising steps to collect attention maps from.
We first note that SD is trained to generate images, while we aim to find text-visual-region correspondences in real images.
At the early reverse diffusion denoising step ($t=T$), an image is random Gaussian image, generating noisy attention maps.
As denoising step $t$ approaches $0$, an image becomes a real image, generating accurate attention maps.
Thus, it is reasonable to start aggregating attention maps after the half of denoising process ($t=T/2$), when an image introduces noise.
Fig.~\ref{fig:timestep} validates our claim in that an aggregated attention map seems noisy when accumulating attention maps from early denoising steps.
Thus, we obtain aggregated cross-attention (ACA) maps $\tilde{\gA}^\text{CA}_{s_r,c_\text{cls}}\in\mathbb{R}^{s_r\times s_r}$ and aggregated self-attention (ASA) maps $\tilde{\gA}^\text{SA}_{s_r,c_\text{cls}}\in\mathbb{R}^{s_r\times s_r\times s_r\times s_r}$  from $t=0$ to $t=T/2$, for each resolution $\{s_r \times  s_r\}^{3}_{r=0}$, where $s_r\in\{8,16,32,64\}$ and $c_\text{cls}$ is the index of a class token in the text prompt.

\subsection{Seed initialization}
\label{sec:seed_generator}
ACA maps $\tilde{\gA}^\text{CA}_{s_r,c_\text{cls}}$ quantify the correspondence between textual features and visual latent features at each attention map coordinate $(i,j)$ for each resolution $s_r$.
We observe that attention tends to disperse as resolution increases, losing the precision of category localization, as shown in Fig.~\ref{fig:ca_from_res}.
Thus, we employ ACA maps at resolutions $8\times8$ and $16\times16$.
In particular, we first extract initial seeds from ACA maps at resolution $8\times8$, $\tilde{\gA}^\text{CA}_{8,c_\text{cls}}$, by gathering spatial coordinates at which ACA value exceeds a threshold $\alpha$:
\begin{equation}
    \mathbb{S}_1 = \{(i,j) \mid \tilde{\gA}^\text{CA}_{8,c_\text{cls}}[i,j]\ge\alpha\}.
\end{equation}
\subsection{Iterative seed generation and spreading}
\label{sec:ses}
While $\mathbb{S}_1$ may contain seeds that point at a query category, they may capture only a small portion of a category and also fail to attend to other multiple instances.
We also note that higher-resolution ASA provides fine-grained attention maps, while lower-resolution ACA provides better localization.
Upon this observation, we use ACA and ASA to perform an iterative seed generation and spreading to gradually refine seeds to higher resolution, obtaining more precise coordinates of seeds.

\begin{figure}
    \centering
    \includegraphics[width=1\linewidth]{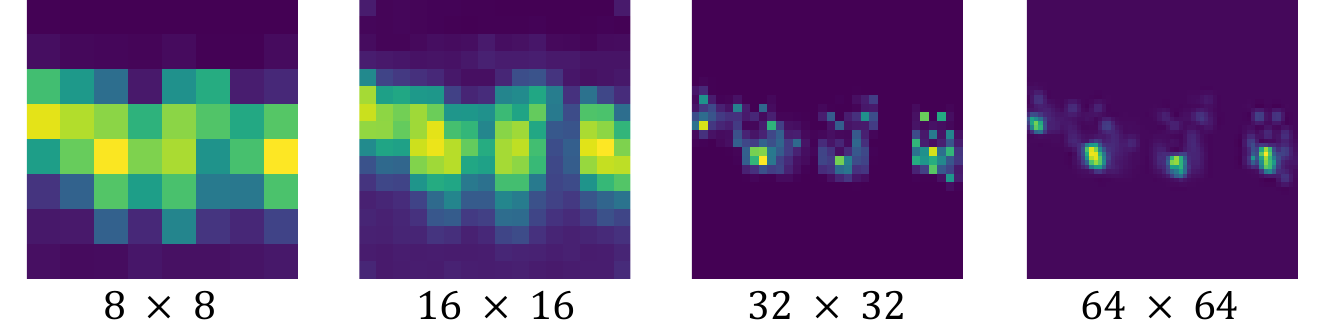}
    \caption{
    Visualization of how aggregated self-attention mask ${\gM}_{s_{k}}$ evolves over the iterations in iterative seed generation and spreading.
    Through the progress, ${\gM}_{s_{k}}$ is shown to exhibit more fine-grained and sparse attention masks.
    Thus, at the end of the iterative seed generation and spreading, we can extract precise seeds at a resolution of $64\times64$, which are then passed to SAM for region expansion.}
    \label{fig:diff_res}
\end{figure}

To this end, we first ``spread'' current extracted seeds $\mathbb{S}_k$ by using ASA maps $\tilde{\gA}^\text{SA}_{s_k,c_\text{cls}}$ to locate features similar to extracted seeds $\mathbb{S}_1$:
\begin{equation}
\label{eq:seed_spread}
    \gM'_{s_k} = \frac{1}{\left|\mathbb{S}_k \right|}\sum_{(i,j)\in\mathbb{S}_{k}}\tilde{\gA}^\text{SA}_{s_k}[i,j,:,:].
\end{equation}
This aggregated attention mask $\gM'_{s_k}$ now attends to visual features that are similarly to current extracted seeds $\mathbb{S}_k$.
Then, $\gM'_{s_k}$ is upsampled via bilinear-upsampling to higher resolution to obtain its higher-resolution counterpart $\gM_{s_{k+1}}\in\mathbb{R}^{s_{k+1}\times s_{k+1}}$:
\begin{equation}
\label{eq:upsampling}
     \gM_{s_{k+1}} = \text{bilinear-upsample}_{s_{k+1}}(\gM'_{s_k}).
\end{equation}
When the resolution is $16\times16$, we further leverage ACA map $\tilde{\gA}^\text{CA}_{16,c_\text{cls}}$ for its better semantic correspondence information, compared to higher resolutions.
To this end, we combine $\gM_{16}$ and  $\tilde{\gA}^\text{CA}_{16,c_\text{cls}}$ via point-wise max operation to capture all visual cues (e.g., seeds):
\begin{equation}
\label{eq:16_ca_map}
    \mathcal{M}_{16} \leftarrow \text{max}(\mathcal{\tilde{A}}_{16}^{\text{CA}} ,\mathcal{M}_{16}).
\end{equation}
Through the above process, the point-wise transformation of the ACA map $\tilde{\gA}^\text{CA}_{16,c_\text{cls}}$ can be observed in Fig.~\ref{fig:diff_res}. 
Again, we extract seeds from $\mathcal{M}_{s_{k+1}}$:
\begin{equation}
\label{eq:seed_extraction}
    \mathbb{S}_{k+1} = \{(i,j) \mid {\gM}_{s_{k+1}}[i,j]\ge\alpha\}.
\end{equation}
Then, we iteratively repeat the process of seed generation and seed spreading using Eq.~\ref{eq:seed_spread},~\ref{eq:upsampling},~\ref{eq:seed_extraction} until the last iteration $K$ that corresponds to the highest resolution of ASA maps ($s_K=64$).

\begin{table*}[!t]
  \centering
  \caption{Comparison with open-vocabulary semantic segmentation models on Pascal VOC~\cite{pascalVOC}, Cityscapes~\cite{Cityscapes}, ADE20K~\cite{Ade20k}, COCO objects~\cite{coco} and Pascal Context~\cite{context}. 
  The performance of other models is reported as in the respective papers.
  Best and second best results are highlighted in \textbf{bold} and \underline{underlined}, respectively.}
  \setlength{\tabcolsep}{.4em}
  \resizebox{\textwidth}{!}{%
  \begin{tabular}{lc c c c c ccccccccccc}
    \toprule
     & & & & & & & & & & \multicolumn{7}{c}{\textbf{mIoU}} \\
    \cmidrule{11-17}
    \textbf{Model} & & \textbf{training-free} & & \textbf{supervision} & & \textbf{preprocess} & & \textbf{SAM} & & VOC-20 & VOC-21 & Cityscapes & ADE & Objects & Pascal59 & Pascal60 \\
    \midrule

    \multicolumn{17}{l}{\textit{Weakly-supervised}} \\
    ReCo~\cite{ReCo} & & \xmark & & \cmark & & \xmark & & \xmark & & 57.7 & 25.1 & 21.1 & 11.2 & 15.7 & 22.3 & 19.9 \\
    GroupViT~\cite{GroupViT} & & \xmark & & \cmark & & \xmark & & \xmark & & 79.7 & 50.4 & 11.1 & 9.2 & 27.5 & 23.4 & 17.7 \\
    MaskCLIP~\cite{maskCLIP} & & \xmark & & \cmark & & \xmark & & \xmark & & 74.9 & 38.8 & 12.6 & 9.8 & 20.6 & 26.4 & 23.6 \\
    TCL~\cite{TCL} & & \xmark & & \cmark & & \xmark & & \xmark & & 83.2 & 55.0 & 24.0 & 17.1 & 31.6 & 33.9 & 30.4 \\
    SAM-CLIP~\cite{SAMCLIP} & & \xmark & & \cmark & & \xmark & & \cmark & & - & 60.6 & - & 17.1 & - & - & 29.2 \\
    \midrule                                 
    
    \multicolumn{17}{l}{\textit{CLIP-based training-free}} \\
    SCLIP~\cite{SCLIP} & & \cmark & & \xmark & & \xmark & & \xmark & & 67.5 & 43.8 & 19.5 & 11.3 & 24.6 & 25.6 & 23.5 \\ 
    ProxyCLIP~\cite{ProxyCLIP} & & \cmark & & \xmark & & \xmark & & \xmark 
    & & {{80.3}} & {{61.2}} & {{38.1}} & {{20.1}} & {{39.1}} & {{39.3}} & {{38.3}} \\ 
    
    CaR~\cite{CaR} & & \cmark & & \xmark & & \xmark & & \xmark 
    & & {{91.6}} & {{67.5}} & {{15.9}} & {{17.8}} & {{37.5}} & {{39.7}} & {{31.5}} \\

    {{NACLIP~\cite{naclip}}} 
        &  & {{\cmark}} 
        &  & {{\xmark}} 
        & & {{\xmark}} 
        & & {{\xmark}} 
        & & {{83.0}} 
        & {{64.1}} 
        & {{38.3}} 
        & {{19.1}} 
        & {{36.2}} 
        & {{38.4}} 
        & {{35.0}} \\

    {{CASS~\cite{CASS}}} 
        &  & {{\cmark}} 
        &  & {{\xmark}} 
        & & {{\xmark}} 
        & & {{\xmark}} 
        & & {{87.8}} 
        & {{65.8}} 
        & {{39.4}} 
        & {{20.4}} 
        & {{37.8}} 
        & {{40.2}} 
        & {{36.7}} \\

    {{CLIPer~\cite{cliper1}}} 
        &  & {{\cmark}} 
        &  & {{\xmark}} 
        & & {{\xmark}} 
        & & {{\xmark}} 
        & & {{89.8}} 
        & {{72.2}} 
        & {{42.5}} 
        & {{25.0}} 
        & {{\underline{44.7}}} 
        & {{\underline{44.6}}} 
        & {{39.5}} \\
    \midrule
    
    \multicolumn{17}{l}{\textit{Diffusion-based training-free}} \\
    OVDiff~\cite{OVdiff} & & \cmark & & \xmark & & \xmark & & \xmark 
    & & {{79.8}} & {{67.4}} & {{17.5}} & {{13.1}} & {{40.6}} & {{30.6}} & {{29.1}} \\

    DiffSegmenter~\cite{DiffSegmenter} & & \cmark & & \xmark & & \xmark & & \xmark & & - & 60.1 & - & - & 37.9 & - & 27.5 \\

    FreeDA~\cite{FreeDA} & & \cmark & & \xmark & & \cmark & & \xmark 
    & & {{86.2}} & {{55.4}} & {{35.9}} & {{23.1}} & {{37.4}} & {{43.4}} & {{39.2}} \\

    {{FreeSeg-Diff~\cite{FreeSeg-Diff}}} 
        &  & {{\cmark}} 
        &  & {{\xmark}} 
        & & {{\xmark}} 
        & & {{\xmark}} 
        & & {{-}} 
        & {{53.3}} 
        & {{-}} 
        & {{-}} 
        & {{31.1}} 
        & {{-}} 
        & {{-}} \\

    {{NERVE~\cite{nerve1}}} 
        &  & {{\cmark}} 
        &  & {{\xmark}} 
        & & {{\xmark}} 
        & & {{\xmark}} 
        & & {{90.1}} 
        & {{69.7}} 
        & {{34.1}} 
        & {{24.0}} 
        & {{43.3}} 
        & {{43.4}} 
        & {{37.7}} 
        \\
    \midrule

    \multicolumn{17}{l}{\textit{SAM-based training-free}} \\
    Grounded-SAM~\cite{GroundedSAM} & & \cmark & & \xmark & & \xmark & & \cmark & & 81.5 & - & 26.9 & 15.3 & 41.5 & 44.0 & - \\
    RIM~\cite{RIM} & & \cmark & & \xmark & & \cmark & & \cmark & & - & \textbf{77.8} & - & - & 44.5 & - & 34.3 \\

    CaR+SAM~\cite{CaR} & & \cmark & & \xmark & & \xmark & & \cmark 
    & & {{\underline{91.7}}} & {{67.9}} & {{16.0}} & {{17.9}} & {{37.7}} & {{40.0}} & {{31.6}} \\

    {{Trident~\cite{shi2024trident}}} 
        &  & {{\cmark}} 
        &  & {{\xmark}} 
        & & {{\xmark}} 
        & & {{\cmark}} 
        & & {{88.7}} 
        & {{70.8}} 
        & {{\textbf{47.6}}} 
        & {{\underline{26.7}}} 
        & {{42.2}} 
        & {{44.3}} 
        & {{\underline{40.1}}} \\
    \midrule

    \rowcolor{Gray}
    \textbf{T2S (Ours)} & & \cmark & & \xmark & & \xmark & & \cmark 
    & & \textbf{92.1} & \underline{74.2} & \underline{47.3} & \textbf{27.1} & \textbf{45.6} & \textbf{47.9} & \textbf{41.7} \\
    \bottomrule
  \end{tabular}
  } 
  \label{tab:main}
\end{table*}

\subsection{Region expansion via SAM}
\label{sec:region_expansion}
In this stage, we aim to employ SAM to perform region expansion: finding a high-quality mask region, pointed by extracted seeds from the previous stage.
SAM is known to give a more precise segmentation mask, when positive point prompts (targets) and negative point prompts (non-targets).
Thus, similar to how we obtain positive point prompts (seeds for categories) in the previous stage, we obtain negative point prompts (seeds for background or other irrelevant features) by simply iteratively extracting seeds and seed spreading, with a small change in Eq.~\ref{eq:seed_extraction}:
\begin{equation}
\label{eq:neg_seeds}
    \mathbb{S}^{\text{neg}}_{k+1} = \{(i,j) \mid \mathcal{M}^{\text{neg}}_{s_{k+1}}[i,j]\le\beta\},
\end{equation}
where $\mathcal{M}^{\text{neg}}_{s_{k+1}}$ is the negative-point-counterpart of $\mathcal{M}_{s_{k+1}}$ obtained with negative seeds from a previous resolution $\mathbb{S}^{\text{neg}}_{k}$.
To extract negative point prompts, we now look for points at which cross-attention values are low.
\begin{figure}
    \centering
    \includegraphics[width=1\linewidth]{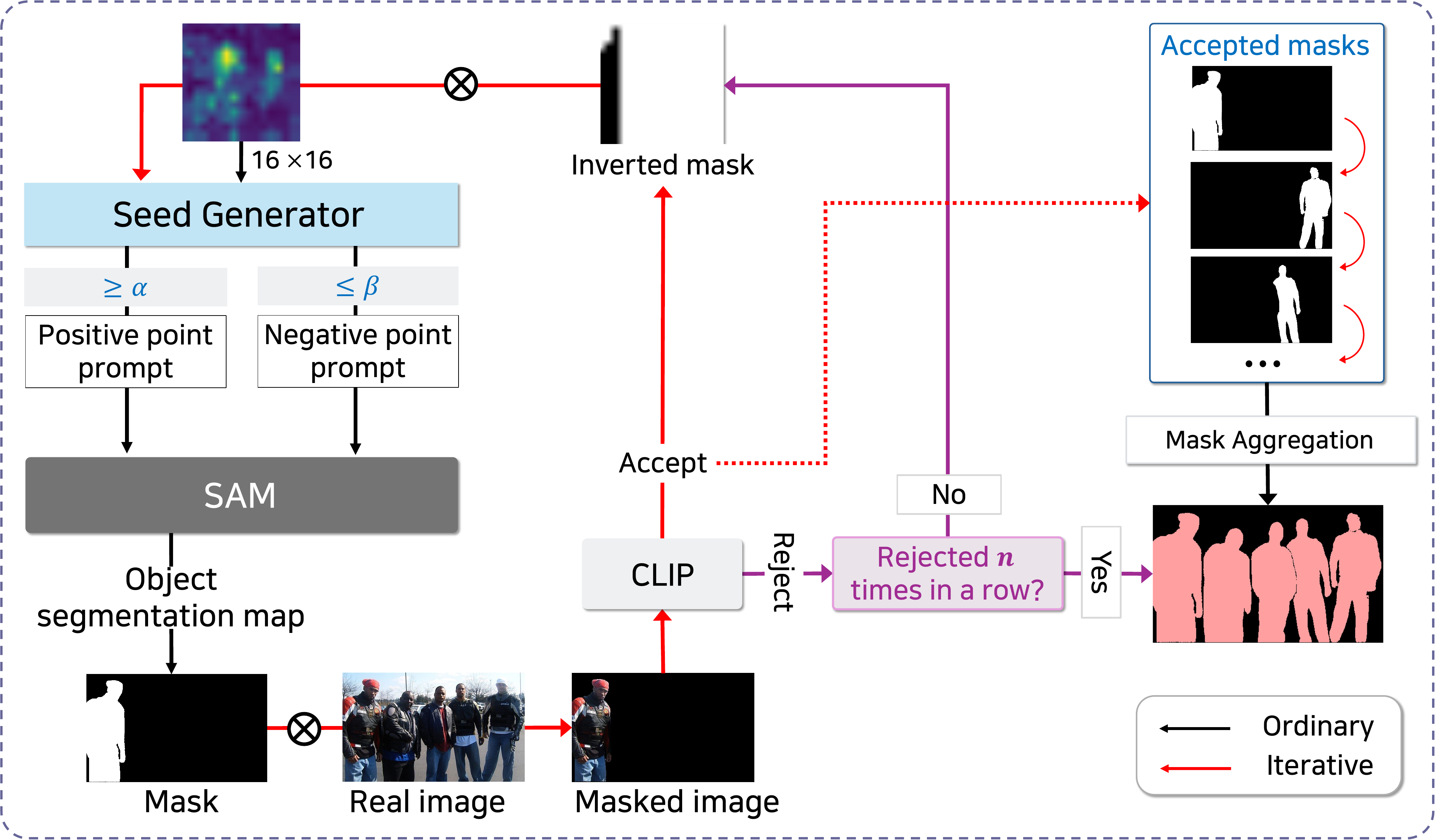}
    \caption{
    Illustration of the process that iteratively aggregates masks before generating the final mask.
    }
    \label{fig:rejection}
\end{figure}

Then, the set of positive seeds $\mathbb{S}_{K}$ and negative seeds $\mathbb{S}^{\text{neg}}_{K}$ extracted at the last iteration $K$ ($\mathbb{S}_{\text{SAM}}=\mathbb{S}_{K} \cup \mathbb{S}_{K}^\text{neg}$) is fed into SAM along with an input image $\vx$ to obtain a segmentation mask $\vm$.
Then, the segmented region is classified by feeding a masked image $\vm\odot\vx$ into CLIP.
Although we start with cross-attention maps that attend to visual attributes of a target category, iterative seed generation and spreading process on small resolutions (i.e., $8\times8, 16\times16, 32\times32, 64\times64$) can lead to false positives.
Thus, we leverage strong zero-shot classification capability of CLIP to filter out such false positives.

\subsection{Iterative seeded semantic segmentation}
\label{sec:iss}
Since cross-attention maps of SDs often seem to struggle with identifying multiple subjects, we propose to iteratively perform seed generation and region expansion from Sec.~\ref{sec:seed_generator} to Sec.~\ref{sec:region_expansion}, followed by masking each found mask region out from $16\times16$ aggregated attention map $\tilde{\gA}^\text{CA}_{16}$, such that seeds will be extracted at different location in each iteration.
We repeat this procedure until CLIP classification rejects segmented regions for more than $n$ ($n=10$ in this work) times in a row or all image regions are covered.
Fig.~\ref{fig:rejection} demonstrates the effectiveness of the iterative seeded semantic segmentation, where masked attention maps become more fine-grained and sparse, providing more accurate localization as the procedure proceeds.

\begin{figure*}
    \centering
    \includegraphics[width=1\linewidth]{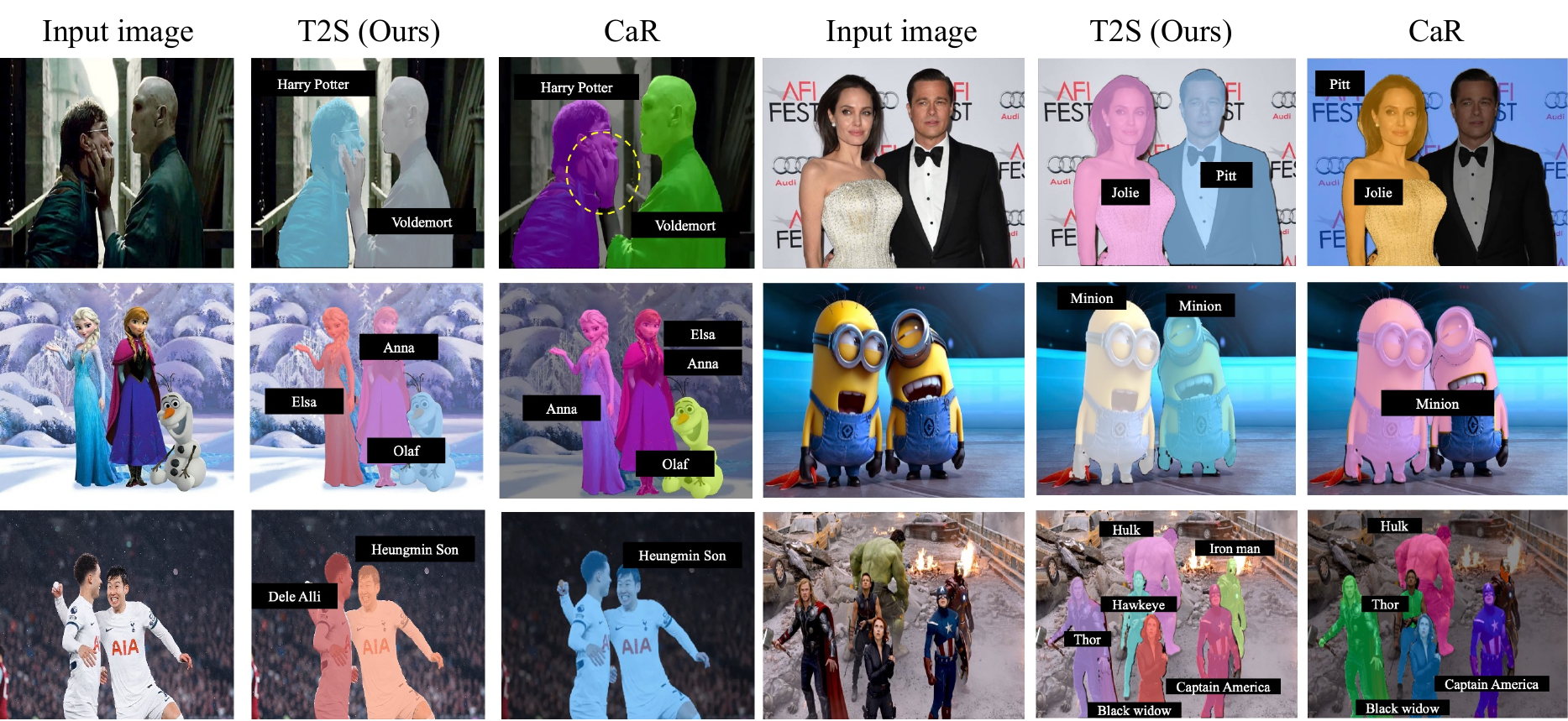}
    \caption{
    Qualitative comparison of T2S with the state-of-the-art method (CaR) under open-vocabulary scenarios}
    \label{fig:openvoca}
\end{figure*}

\begin{figure*}
    \centering
    \includegraphics[width=1\linewidth]{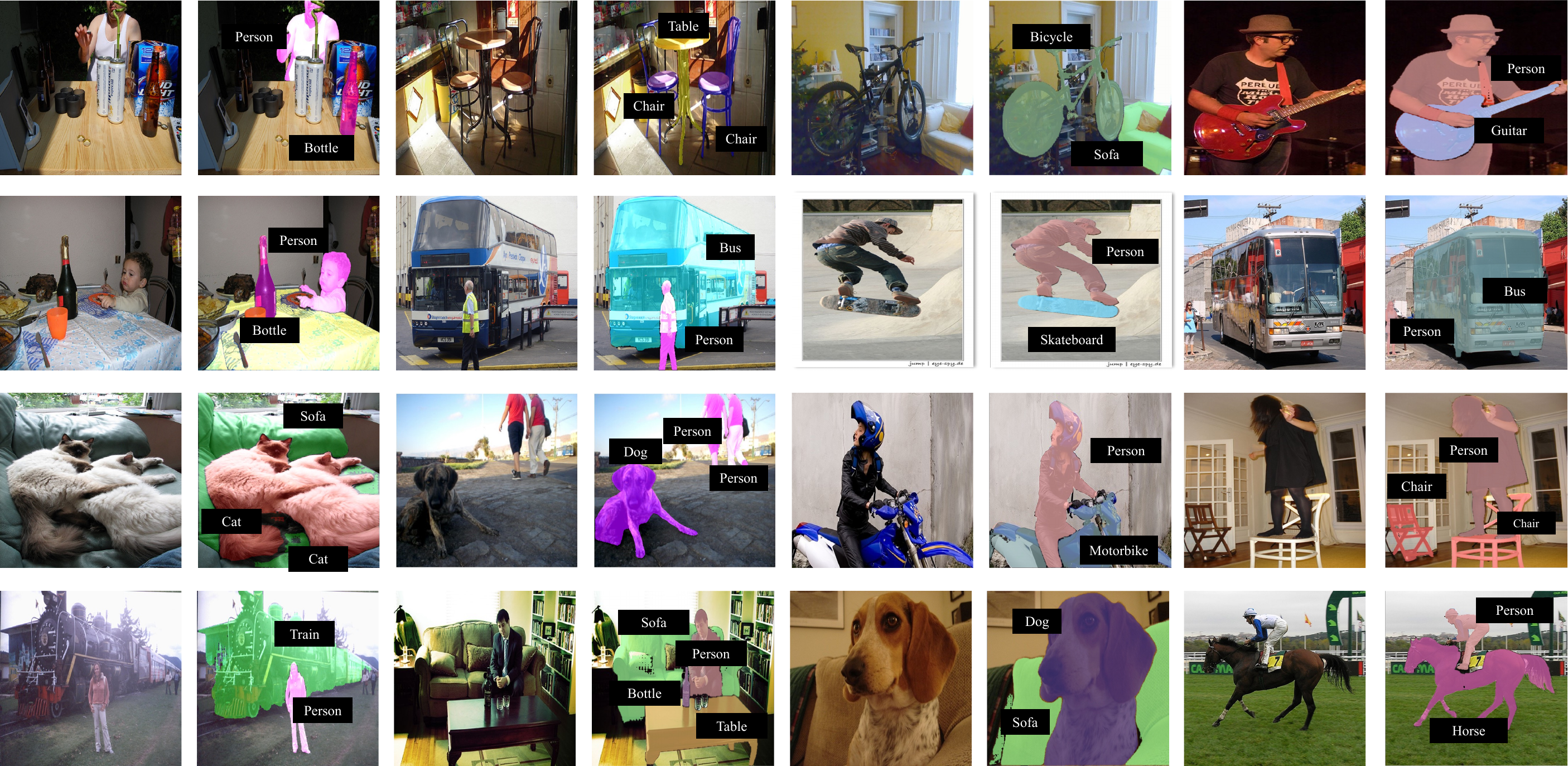}
    \caption{
    Qualitative results of T2S under multi subject scenarios.
    }
    \label{fig:multi}
\end{figure*}

\begin{table}[t]
  \centering
    \caption{
  Ablation on our proposed module. Init denotes the Initialization of Seed Generator, SG denotes Seed Generator, ISSS denotes Iterative Seeded Semantic Segmentation from our method. 
  All experiments are conducted on Pascal VOC 20.
  }
  \setlength{\tabcolsep}{0.3em}
  \begin{tabular}{@{\hspace{10pt}}c@{\hspace{15pt}}c@{\hspace{15pt}}c@{\hspace{15pt}}c@{\hspace{15pt}}c}
    \toprule
    \cellcolor{white}\textbf{Init} & \multicolumn{2}{c}{\textbf{SE}} & \textbf{ISSS} & \textbf{VOC} \\
    \cmidrule{2-3}
    \shortstack{Token\\weight} & \shortstack{Seed\\generation} & \shortstack{Negative\\seed} & \shortstack{Iterative\\SSS} & mIoU\\
    \midrule
    - & \cmark & - & - & 73.1 \\
    \cmark & \cmark & - & - & 76.8 \\
    \cmark & \cmark & \cmark & - & 82.3 \\
    \cmark & \cmark & \cmark & \cmark & 92.1 \\
    \bottomrule
  \end{tabular}
  \label{tab:ablation_module}
\end{table}

\section{Experiments}
\label{sec:experiments}
\subsection{Dataset and Evaluation Metric} We evaluate our method Text-to-Seed (T2S) on common benchmarks: Pascal VOC 2012~\cite{pascalVOC}, Pascal Context~\cite{context}, Cityscapes~\cite{Cityscapes}, COCO object~\cite{coco}, ADE20K~\cite{Ade20k}.
Following previous works~\cite{OVdiff}, we consider the background class in Pascal VOC 2012 and Pascal Context dataset.
\textbf{Pascal VOC 2012} is consist of 1449 images with 20 classes for VOC-20 and 21 classes for VOC-21.
\textbf{Pascal Context} comprises 5104 images with 59 classes for Pascal59 and 60 classes for Pascal60.
\textbf{Cityscapes} includes 500 images with 19 classes without background class.
\textbf{COCO object} includes 5000 images with 80 classes.
\textbf{ADE20k} comprises 2000 images with 150 classes without background class. 
Following standard evaluation protocol, the mean Intersection-over-Union (mIoU) metric is used to evaluate the models.

\subsection{Implementation Details} We utilize the Stable Diffusion v1.4 to extract CA map and SA map conditioned on input text prompt.
We utilize the text prompt `A photo of \texttt{\textless class\textgreater}' to condition the SD model when extracting the CA map.
For region classification, we utilize the CLIP model with ViT-L/14 as a backbone, as in previous works~\cite{CaR, ProxyCLIP}.
We use the seed generation threshold parameter $\alpha$ as 0.5 and $\beta$ as 0.3 to extract the seeds from initialization map at each resolution of the SD model.
All experiments are performed on NVIDIA RTX 4090.

\subsection{Quantitative results} 
Tab.~\ref{tab:main} reports the performance of recent OVSS methods and our method, T2S across several datasets, namely VOC-20, VOC-21, Cityscapes, ADE, Objects, Pascal59, and Pascal60.
All the results of other methods are taken from the numbers reported in the respective papers, except for Grounded-SAM, which is reported in~\cite{mess}.
The results demonstrate the outstanding performance of our method T2S, compared to previous works that employ CLIP, Diffusion, SAM, or the combination.

While the original paper of SAM~\cite{SAM} suggests that SAM supports the text prompt, the official implementation of SAM does not support the text prompt.
GroundedSAM~\cite{GroundedSAM} provides a simple workaround to the lack of the semantic modeling capability of SAM.
However, GroundedSAM relies on the bounding box produced by Grounding DINO, which is trained on annotated bounding boxes.
Another work~\cite{SAMCLIP} also fine-tunes SAM and CLIP with mask annotation for OVSS.
The reliance on such rich annotation exposes the risks of poor performance on unseen categories.
In parallel, other works employ SAM to refine the initial masks produced by CLIP~\cite{CaR} or Stable Diffusion~\cite{RIM}.
Such frameworks rely on the quality of initial masks, unable to fully exploit the capabilities of SAM.

On the other hand, our proposed framework, T2S, better exploits the spatial coherence capabilities of SAM by utilizing it as region expansion.
Furthermore, in contrast to previous works that employ SD for initial mask generation, T2S utilizes SD for seed initialization (i.e., object localization), which is less susceptible to noise and errors, compared to the whole masks.

\subsection{Ablation Study} 
We conduct ablation studies to evaluate the effectiveness of the various modules introduced in our methodology.
Ablation study experiments are performed on Pascal VOC 2012 with mean Intersection over Union (mIoU) as the primary metric to quantitatively assess the influence of each design choice and module.

\begin{table}[t]
\centering
\caption{Ablation study on the selection of the timestep at which to initiate the denoising process.}
\setlength{\tabcolsep}{6pt}
\renewcommand{\arraystretch}{1.1}
\begin{tabular}{@{\hspace{20pt}}c@{\hspace{20pt}}c@{\hspace{20pt}}c@{\hspace{20pt}}c@{\hspace{20pt}}}
\hline
 & \textbf{3/4 T} & \textbf{1/2 T} & \textbf{1/4 T} \\
\hline
\hline
\textbf{mIoU} & 90.4 & \textbf{92.1} & 89.7 \\
\hline
\end{tabular}
\label{tab:ablation_timestep}
\end{table}

\begin{table}[t]
\centering
\caption{mIoU by {aggregated cross-attention (ACA) map resolutions used for seed initialization} in SGS.}
\setlength{\tabcolsep}{6pt}
\renewcommand{\arraystretch}{1.15}
\centering
\begin{tabular}{@{\hspace{20pt}}c @{\hspace{50pt}} r @{\hspace{40pt}}}
\toprule
\textbf{Resolutions} & \textbf{mIoU} \\
\midrule
\midrule
8 & 88.2 \\
8, 16 & \textbf{92.1} \\
8, 16, 32 & 82.4 \\
8, 16, 32, 64 & 76.3 \\
\bottomrule
\end{tabular}
\label{tab:res}
\end{table}

\begin{table}[t]
\centering
\caption{mIoU by seed generation threshold $\alpha$.}
\begin{tabular}{@{\hspace{20pt}}c @{\hspace{50pt}} r @{\hspace{40pt}}}
\toprule
$\alpha$ & \textbf{mIoU} \\
\midrule \midrule
0.3 & 87.2 \\
0.4 & 91.7 \\
0.5 & \textbf{92.1} \\
0.6 & 90.3 \\
0.7 & 84.2 \\
\bottomrule
\end{tabular}
\label{tab:alpha}
\end{table}
\begin{table}[t]
\centering
\caption{
Ablation study comparing fixed iteration counts ($R$) and our adaptive rejection-based strategy using CLIP.}
\setlength{\tabcolsep}{5pt}  
\renewcommand{\arraystretch}{1.1}  
\begin{tabular}{@{\hspace{20pt}}c@{\hspace{20pt}}c@{\hspace{20pt}}c@{\hspace{20pt}}c@{\hspace{20pt}}}
\hline
 & \textbf{$R=1$} & \textbf{$R=5$} & \textbf{Ours} \\
\hline
\hline
\textbf{mIoU} & 82.3 & 86.8 & 92.1 \\
\hline
\end{tabular}
\label{tab:ablation_rejectcount}
\end{table}

\noindent\textbf{Module Ablation.} 
Tab.~\ref{tab:ablation_module} presents the results of an ablation study evaluating the effectiveness of each module: namely, token weighting in the initialization of seed generator; negative seed point prompt; and iterative procedure of seed generation and spread. 
Using seed generation with SAM for mask generation as the baseline model, we progressively apply each module and observe the performance difference.
Our basic pipeline generates point prompts from the aggregated attention maps of SD, which can be considered as applying only seed generation from Seed Generator.
Then, to get more precise point prompts, we utilize token weighting process from Initialization of Seed Generator, which has indeed led to performance improvement.
Extracting and feeding negative point prompts to T2S has led to substantial performance improvement.
Introducing iterative procedure to better handle multiple objects has been observed to provide even more performance improvement.

\noindent\textbf{Ablation on hyper-parameters.} 
Tab.~\ref{tab:ablation_timestep} presents the ablation study on the starting diffusion denoising step to extract aggregated attention maps.
The first row of the table indicates the timestep at which the denoising process begins, with $T$ set to 75 steps. 
We observe that initiating the denoising from the midpoint, at $1/2T$, leads to the best performance as it allows for greater focus on the object.
This is because images become close to Gaussian distribution and hence noisier, as the denoising is performed at earlier steps.

Tab.~\ref{tab:res} illustrates the effect of the resolution of attention maps.
Using higher resolution of attention maps for seed generation initialization leads to a decrease in performance.
This is because attention disperses at higher resolution, losing the category localization information, as discussed in Sec.~\ref{sec:initialization} and illustrated in Fig.~\ref{fig:ca_from_res}.

\begin{table}[t]
\centering
 \caption{Generalization performance on ISPRS Potsdam dataset.}
\resizebox{0.8\columnwidth}{!}{
\begin{tabular}{@{\hspace{20pt}} c @{\hspace{20pt}} c @{\hspace{20pt}} c}
			\bottomrule
               & \textbf{CaR+SAM} &  \textbf{T2S (Ours)}   \\
              \bottomrule
               \textbf{mIoU} & 81.3 & 83.2  \\

			\bottomrule
	\end{tabular}}
\label{tab:generalization}
\end{table}

\begin{table}
    \centering
    \caption{
    {\textbf{Computational cost of T2S.}
    Runtime, GPU memory, and model-call counts are reported per image,
    while the average number of iterations is reported per queried class.
    CLIP calls correspond to mask verification.}
    }
    \label{tab:computational_cost}
    \small
    \setlength{\tabcolsep}{7pt}
    \renewcommand{\arraystretch}{1.1}
    \begin{tabular}{@{}lc@{}}
        \toprule
        Metric & Value \\
        \midrule
        Inference time
            & 188 s/image (3.13 min) \\
        Peak GPU memory
            & $\sim$20 GB \\
        Average number of iterations
            & 9.1/class \\
        SAM image-encoder calls
            & 1/image \\
        SAM mask-decoder calls
            & 311.7/image \\
        CLIP mask-verification calls
            & 21.8/image \\
        \bottomrule
    \end{tabular}
\end{table}

\begin{table}
\centering
 \caption{{The analysis of computational complexity with respect to the number of objects, measured in GFLOPs.
 }}
    \resizebox{0.9\columnwidth}{!}{
    \begin{tabular}{ccccc}
			\bottomrule
			   & \textbf{$n=1$} & \textbf{$n=2$} &  \textbf{$n=3$} &  \textbf{$n > 5$}  \\
			\bottomrule
			  \textbf{Seed Generator} & 3623 & 3630 & 3618 & 3651 \\
			\bottomrule
	\end{tabular}}
\label{tab:computational cost by modules}
\end{table}
\begin{table}[t]
\centering
\caption{Peak GPU memory and computational-cost comparison with representative training-free OVSS baselines on VOC-20. Memory usage is reported in GB, and computational cost is reported in GFLOPs.
}
\resizebox{0.6\columnwidth}{!}{
\begin{tabular}{lcc}
\toprule
 & GPU Memory &  GFLOPs \\
\midrule
CaR~\cite{CaR}  &  ~8.6 GB & 338\\
FreeDA~\cite{FreeDA} & ~23.4 GB & 588\\
OVDiff~\cite{OVdiff}  & ~21.0 GB   &1492\\
T2S (Ours)  & ~20.0 GB & 3851\\

\bottomrule
\end{tabular}}
\label{tab:time per img}
\end{table}

Tab.~\ref{tab:alpha} reports the effect of seed generation threshold parameter $\alpha$.
Without extensive hyperparameter search, $\alpha=0.5$ is shown to provide the best performance.
Tab.~\ref{tab:ablation_rejectcount} represents an ablation study on the number of iterations $R$ during Iterative Seeded Semantic Segmentation. 
We evaluate the performance when the number of iterations $R$ is fixed to either $1$ and $5$ and compare it with our final version, where the number of iterations $R$ is dynamically adjusted instance-wise using the rejection count $n$ ($n=10$ in our work).
In other words, T2S iteratively extracts masks until the number of rejections by CLIP reaches $n=10$.
The outstanding performance by T2S compared to other variants suggests the importance of iteratively collecting masks.
This is because aggregated cross-attention maps provide precise localization of one object at a time.
The ablation study results validate that we have overcome such limitation of Stable Diffusion via iterative seed generation and expansion.
However, the iterative procedure does not require the multiple passes of SD for each category, as we perform seed generation and expansion on aggregated attention maps, which need to be extracted from SD only once for each category.
{
Tab.~\ref{tab:computational_cost} reports the practical inference cost of T2S. T2S requires 188\,s per image and approximately 20\,GB of peak GPU memory, with an average of 9.1 iterations per queried class.
As shown in Tab.~\ref{tab:computational cost by modules}, the seed-generation cost varies by less than 1\% across the evaluated object-count bins (3.62--3.65\,TFLOPs), indicating that this stage is largely insensitive to the number of objects.} {Tab.~\ref{tab:time per img} further compares the peak GPU memory usage and computational cost (GFLOPs) of T2S with those of representative training-free OVSS baselines on VOC-20. T2S maintains peak GPU memory usage comparable to that of diffusion-based baselines while incurring a higher computational cost than CaR. This overhead mainly arises from repeated SAM mask decoding and CLIP-based candidate verification, and it remains a practical limitation. Nevertheless, these iterative operations enable the discovery of multiple target instances and the rejection of false-positive candidates, both of which are integral to the T2S pipeline.}

\subsection{Qualitative Results}
\begin{figure}
    \centering
    \includegraphics[width=1\linewidth]{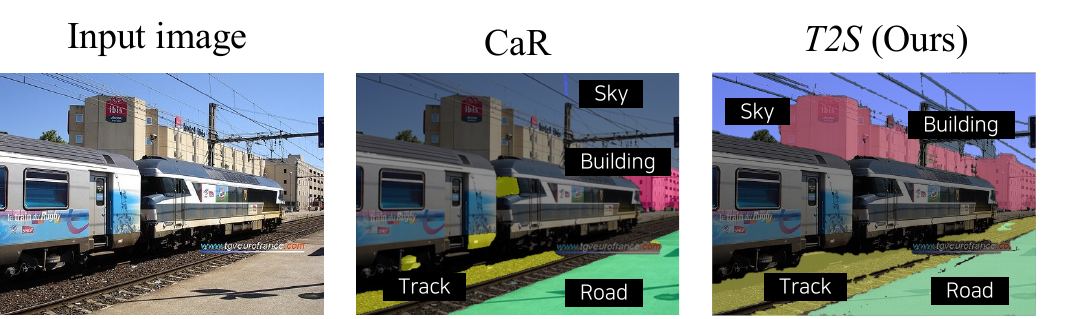}
    \caption{Qualitative results of our method on background classes in the Pascal59 dataset.
    }
    \label{fig:background}
\end{figure}

\begin{figure}
    \centering
    \includegraphics[width=1\linewidth]{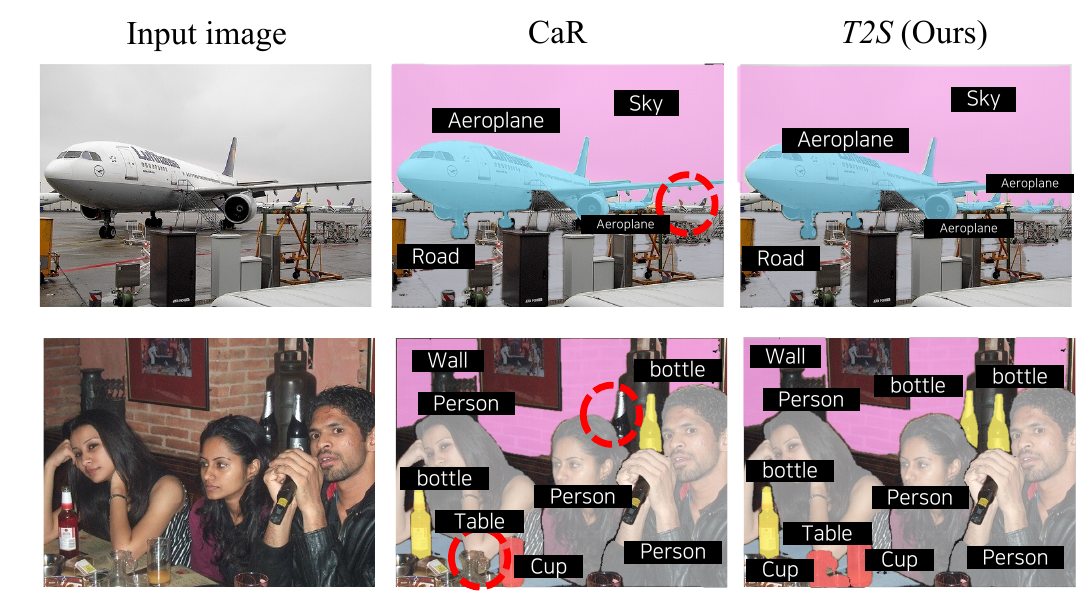}
    \caption{Qualitative comparison of T2S and CaR on small objects and occlusion.
    }
    \label{fig:compare_with_CaR}
\end{figure}

\begin{figure*}
    \centering
    \includegraphics[width=1\linewidth]{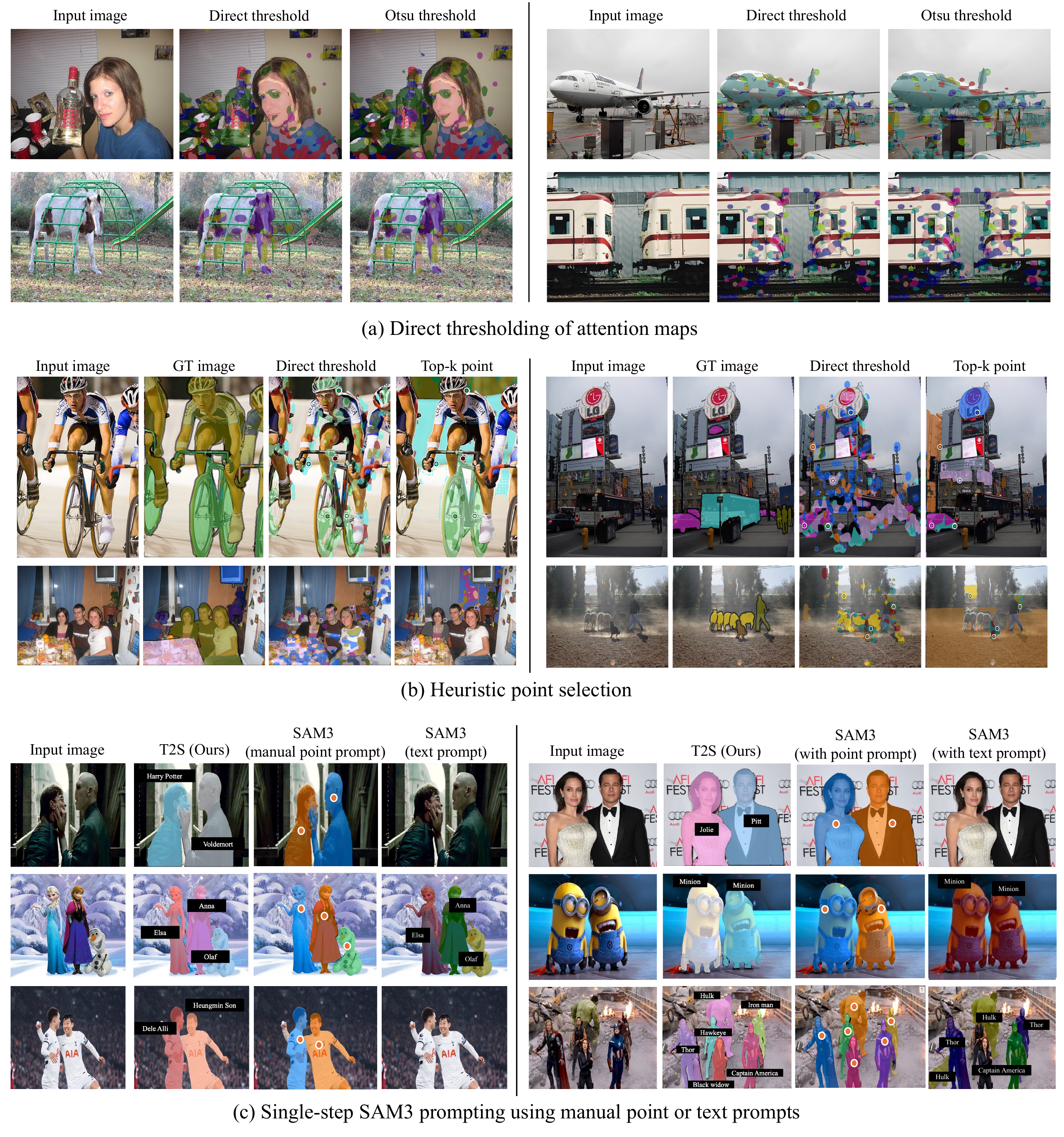}
    \caption{{Qualitative comparison of T2S with simpler alternatives.
    }}
    \label{fig:formulation}
\end{figure*}

Fig.~\ref{fig:multi} illustrates the qualitative results of T2S under multi-object scenarios.
{These results show that T2S can produce high-quality masks in multi-object scenes, despite the known limitation of Stable Diffusion in modeling multiple objects.}
The results validate the effectiveness of our proposed iterative seeded semantic segmentation procedure in overcoming the limitation.

Meanwhile, Fig.~\ref{fig:openvoca} exhibits the qualitative results of T2S under Open-Vocabulary Semantic Segmentation (OVSS) scenarios.
The high-quality masks produced by T2S for arbitrary categories, particularly fictional characters, validate the effectiveness of employing Stable Diffusion to localize objects while using SAM to perform region expansion.
It would be difficult to obtain the high-quality masks for such arbitrary categories, if SAM or other components are fine-tuned on a fixed number of categories with bounding box or mask annotations as in~\cite{SAMCLIP,GroundedSAM}.
On the other hand, T2S exploits the text-spatial-image-feature alignment capability of SD and the spatial coherence modeling ability of SAM in an off-the-shelf manner, leading to high-quality masks for arbitrary categories.
In particular, SD is trained on a large collection of web-crawled image-caption pairs, while SAM is trained to segment general objects without semantic understanding.
Thus, SD enables T2S to obtain an accurate localization of arbitrary objects, while SAM helps T2S segment localized categories.

T2S further demonstrates robust mask generation under complex visual conditions, including intricate background structures (Fig.~\ref{fig:background}) and small occluded objects (Fig.~\ref{fig:compare_with_CaR}).
Specifically, Fig.~\ref{fig:compare_with_CaR} shows that T2S recovers precise boundaries for small occluded targets, whereas CaR misses several small objects as highlighted by dotted red circles.

Fig.~\ref{fig:formulation} further compares T2S with simpler alternatives that use the same underlying components without the proposed seed-generation-and-spreading formulation.
Directly thresholding diffusion attention maps often produces incomplete or noisy masks, since attention maps provide localization cues rather than full object regions.
Selecting top-$k$ attention responses as point prompts improves boundary quality through SAM, but remains vulnerable to incorrect attention peaks and missed instances.
Direct SAM3 prompting can generate accurate masks when reliable manual or text prompts are available, but such prompts are difficult to obtain for arbitrary named entities and multi-instance scenes.
In contrast, T2S automatically derives text-conditioned seed prompts through CA-based localization and ASA-based seed spreading, enabling SAM to perform region expansion without manual prompts or coarse mask inputs.

\begin{figure}
    \centering
    \includegraphics[width=1\linewidth]{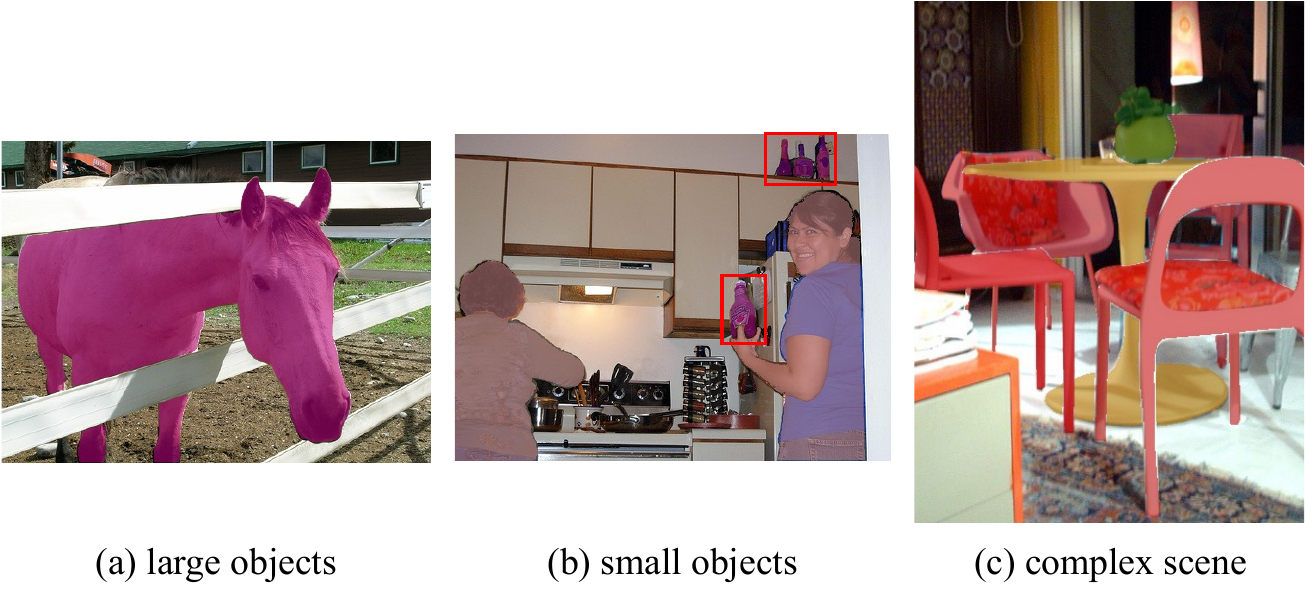}
    \caption{{Qualitative analysis across object scales and scene complexity.}}
    \label{fig:object_scene}
\end{figure}

\begin{figure}
    \centering
    \includegraphics[width=1\linewidth]{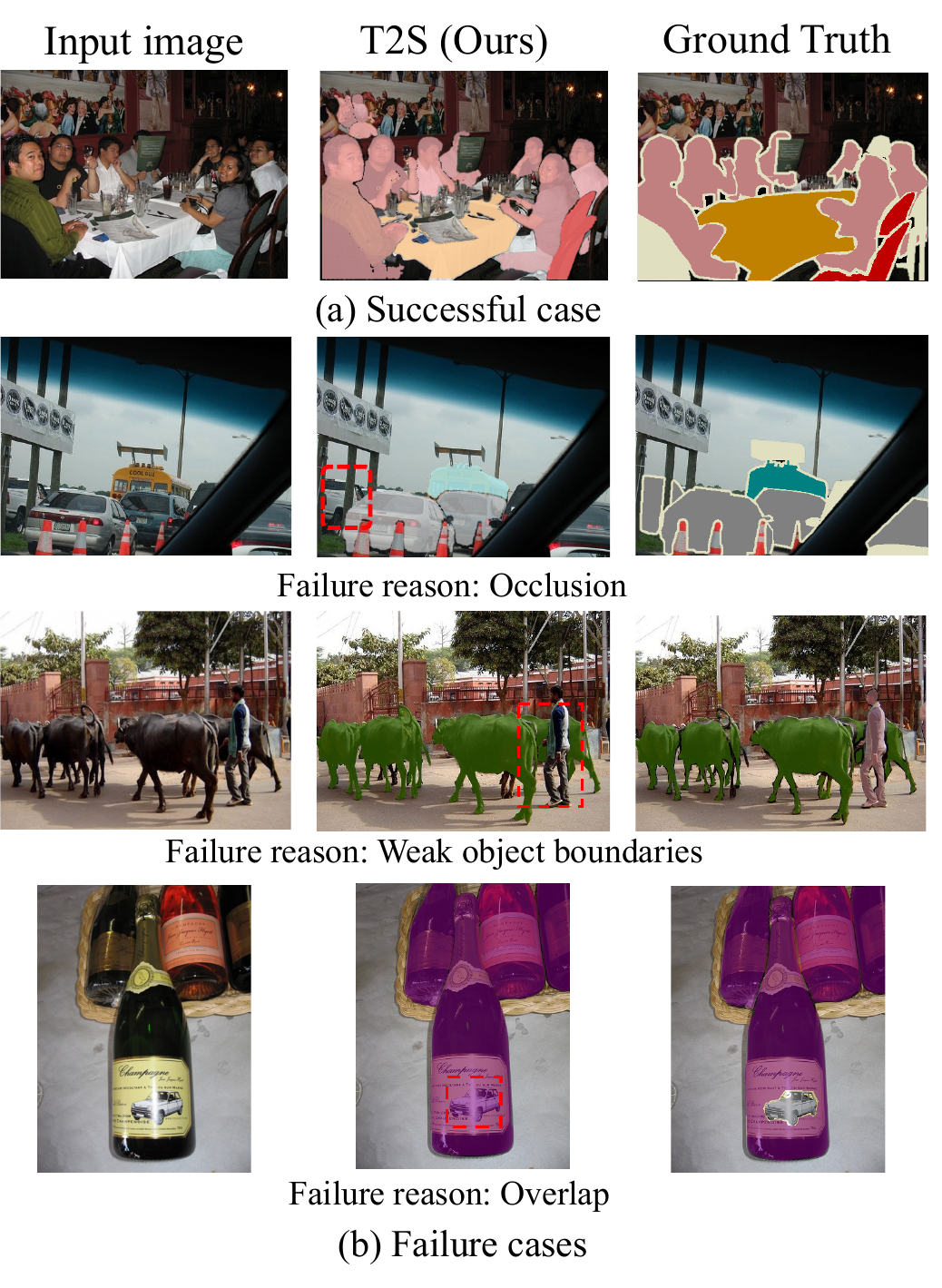}
    \caption{{\textbf{Qualitative results with failure analysis}.
    (a) Successful case shows the very fine-grained segmentation output.
    (b) The failure cases show the object missed by T2S (highlighted by a dotted red box).
    T2S may fail in scenes with severe occlusion, overlapping target and non-target regions, or weak object boundaries. These cases can lead to missed instances or leakage into nearby objects.
    }}
    \label{fig:failure}
\end{figure}

Fig.~\ref{fig:object_scene} complements Figs.~\ref{fig:background} and~\ref{fig:compare_with_CaR} by providing qualitative evaluations across varying object scales and scene complexities.
These visual comparisons confirm the overall robustness and accuracy of T2S.
Fig.~\ref{fig:failure} illustrates potential failure modes, indicating that T2S may encounter difficulties under severe occlusion, faint object boundaries, or target/non-target spatial overlaps, leading to unsegmented regions or boundary leakage into adjacent objects and background regions.

\label{sec:additional experimental analyses}

\subsection{Evaluation on Out-of-Domain images}
To further stress-test the generalization capability of T2S, we evaluate T2S and one of the recent state-of-the-art models, CaR+SAM~\cite{CaR}, on aerial and satellite images, namely the ISPRS Potsdam~\cite{ISPRSdataset} dataset.
Such aerial and satellite images substantially differ from the datasets used for training SAM or Diffusion models.
The results reported in Tab.~\ref{tab:generalization} show that T2S outperforms CaR+SAM, which is one of the recent state-of-the-art models, demonstrating the strong generalization capability of our proposed framework.
\section{Conclusion}
In this work, we introduce Text-to-Seed (T2S), a training-free approach that recasts to open-vocabulary semantic segmentation as text-guided seed localization and seed-based region expansion. 
By using Stable Diffusion to generate attention-based seed points and SAM to expand them into full object masks, T2S avoids the limitations of coarse-mask refinement while effectively combining semantic grounding with spatial segmentation. 
Results on standard OVSS benchmarks show that the proposed framework achieves strong performance without additional training or annotations. 
Overall, our findings demonstrate the promise of using SAM as a seed-driven expansion model and suggest a practical direction for more flexible open-vocabulary segmentation.

\nocite{*}
\section{Acknowledgments}
This work was supported by the Institute of Information and communications Technology Planning and evaluation (IITP) grants (No.RS-2025-25422680, No. RS-2020-II201373) and the National Research Foundation of Korea (NRF) grant funded by the Korea government (MSIT) (No. RS-2025-24533064).

{
    \small
    \bibliographystyle{IEEEtranN}
    \bibliography{main}
}


\end{document}